# Human-centric Transfer Learning Explanation via Knowledge Graph
## [Extended Abstract]

Yuxia Geng[1], Jiaoyan Chen[2], Ernesto Jiménez-Ruiz[3,4], Huajun Chen[1,5]

[1]College of Computer Science, Zhejiang University, China
[2]Department of Computer Science, University of Oxford, UK
[3]The Alan Turing Institute, London, UK
[4]Department of Informatics, University of Oslo, Norway
[5]ZJU-Alibaba Joint Lab on Knowledge Engine, China

## Abstract

Transfer learning which aims at utilizing knowledge learned from one problem (source domain) to solve another different but related problem (target domain) has attracted wide research attentions. However, the current transfer learning methods are mostly uninterpretable, especially to people without ML expertise. In this extended abstract, we brief introduce two knowledge graph (KG) based frameworks towards human understandable transfer learning explanation. The first one explains the transferability of features learned by Convolutional Neural Network (CNN) from one domain to another through pre-training and fine-tuning, while the second justifies the model of a target domain predicted by models from multiple source domains in zero-shot learning (ZSL). Both methods utilize KG and its reasoning capability to provide rich and human understandable explanations to the transfer procedure.

## Introduction

Machine learning (ML) systems such as deep neural networks are becoming increasingly complex for more power. It costs big data and much time for training. Transfer learning refers to a series of ML methods that aim to utilize samples, models or model parameters gained while solving one problem (source domain) to solve another related but different problem (target domain) (Pan *et al.* 2010). It now has attracted wide research attentions in reusing learned models in new contexts, dealing with training data shortage, etc.

Transfer learning has shown its effectiveness and efficiency in many applications such as using satellite image features to predict global poverty (Jean *et al.* 2016). However, an arbitrary transfer might decrease the performance in the target domain, which is called a negative transfer. Some work analyze the theoretical or empirical principles to guide the transfer, but do not explain a specific positive or negative transfer. Some other work develop methods to transform the samples or model parameters from source domains towards a target domain, but they are often black-box models.

On the other hand, ML explanation, including model interpretation and prediction justification is now becoming a more and more urgent requirement, especially in making critical or safety-sensitive decisions. Among them, human-centric explanation, which is enriched by human understandable common sensne and background knowledge (Biran and McKeown 2017), is especially important for none ML experts to optimize a learning system or understand the prediction. However, there are currently few human-centric explanation studies for transfer learning problems.

Knowledge graph (KG), which originally refers to a knowledge base used by Google to enrich its searching results with different data sources, now often means a wide range of semantic data including RDF (Resource Description Framework) graph, OWL (Web Ontology Language) ontology, Linked Data and so on. It's widely used to model common sense or background knowledge with the potential of knowledge representation and reasoning. In this paper, we introduce our idea and framework that utilize KGs to support human-centric explanation of transfer learning with two case studies. The first case study is pair-wise feature transferability explanation. We infer evidences from explicit ontology axioms (facts) to understand the transferability of features learned by Convolutional Neural Networks (CNNs). The second case study is **z**ero-**s**hot **l**earning e**X**planation (X-ZSL). We use external knowledge to justify the CNN of a new class (i.e., class unseen in the training data) predicted by a Graph Convolutional Network (GCN) with CNNs of training classes.

## Feature Transferability Explanation

We assume features learned by a CNN is transferred from a source domain $\mathcal{D}_\alpha$ to a target domain $\mathcal{D}_\beta$, where $\mathcal{D}_\alpha$ has a large training sample set while $\mathcal{D}_\beta$ has a small training sample set. A feature transfer from $\mathcal{D}_\alpha$ to $\mathcal{D}_\beta$, denoted as $\mathcal{F}_{\alpha\rightarrow\beta}$ includes three steps: *(i)* pre-train a CNN with samples from $\mathcal{D}_\alpha$, *(ii)* transfer (reuse) the parameters from the input layer to a specific convolutional layer, which corresponds to a specific feature, and *(iii)* fine tune the CNN with samples from $\mathcal{D}_\beta$. The model trained with $\mathcal{F}_{\alpha\rightarrow\beta}$ is denoted as $\mathcal{M}_{\alpha\rightarrow\beta}$, while the model trained with samples in $\mathcal{D}_\beta$ alone is denoted as $\mathcal{M}_\beta$. $\mathcal{F}_{\alpha\rightarrow\beta}$ is defined as a *positive transfer* if $\mathcal{M}_{\alpha\rightarrow\beta}$ outperforms $\mathcal{M}_\beta$ on the testing sample set from $\mathcal{D}_\beta$, and a *negative transfer* otherwise.

We use OWL ontologies (expressive KGs) to model the background knowledge and data for human-centric explanation. In particular, each domain has an ontology composed of a terminology axiom set $\mathcal{T}$ and an assertion axiom set $\mathcal{A}$. $\mathcal{T}$ models the background knowledge while

$\mathcal{A}$ models the data. Each ontology axiom is equal to a fact of the domain. Considering a case of flight departure delay forecasting in US, and a domain identified by the carrier Delta Airline (DL), the original airport LAX and the target airport ORD, the domain's terminology axioms include concept definitions e.g., $Carrier \sqcap \exists hasStockPrice.Float \sqsubseteq ListCarrier$, relation definitions e.g., $hasCarrier \circ hasCarHub \sqsubseteq hasDepHub$, etc., while its assertion axioms include individuals e.g., $LAX$, $DL$ and $dep$ (representing a flight departure), individual classifications e.g., $Airport(LAX)$, individual relations $locatedIn(LAX, CA)$, etc. Such an ontology has the capability of reasoning, namely inferring underlying axioms e.g., $hasDepHub(dep, LAX)$, and can be aligned with external KGs for knowledge enrichment.

We use *explanatory evidences* to interpret the transferability of features. An explanatory evidence is an ontology axiom or a combination of ontology axioms whose co-existence in the source and target domains is highly correlated with the transferability of a feature. We aim at discovering a set of positive (negative) explanatory evidences that can explain the positive (negative) transferability of a specific transfer (cf. Example 1).

**Example 1.** *(Explanatory Evidence)* In the above case of flight departure delay forecastsing, considering three learning domains: $\mathcal{D}_{(DL,ORD,LAX)}$ for Delta Airlines from ORD to LAX, $\mathcal{D}_{(B6,LAX,JFK)}$ for JetBlue from LAX to JFK and $\mathcal{D}_{(AA,ORD,SFO)}$ for American Airlines from ORD to SFO, for a specific CNN feature, let axiom combination "$locatedIn(ori, East)$" and "$ListCarrier(car)$" be an example of positive explanatory evidence, let axiom combination "$hasOri(dep, ORD)$" and "$locatedIn(des, CA)$" be an example of negative explanatory evidence, human beings can understand the feature's transferability as "the transfer from $\mathcal{D}_{(DL,ORD,LAX)}$ to $\mathcal{D}_{(AA,ORD,SFO)}$ is positive as the original airport of both source and target domains is located in the east part of US and the carrier of both is list airline company" and "the transfer from $\mathcal{D}_{(DL,ORD,LAX)}$ to $\mathcal{D}_{(B6,LAX,JFK)}$ is negative as the original airport of both source and target domains is ORD and the destination airport of both is located in California".

**Framework and Technical Challenges**

Figure 1 presents the computation framework for explanatory evidences of feature transferability. It mainly includes two parts: *(i)* knowledge construction and enrichment, and *(ii)* correlative reasoning for evidence inference.

Knowledge construction refers to the modeling of background knowledge and data of a domain with an ontology, while knowledge enrichment refers to align individuals of initial ontology with instances of external KGs like DBPedia so as to importing more background and common sense axioms. This part finally leads to a complete ontology mentioned above. For example, axioms $hasDepHub(dep, LAX)$ and $ListCarrier(DL)$ depend on some external knowledge about DL's stock price and DL's relationship with LAX.

In this part, one technical challenge is selecting a part of indvduals (called *root individuals*) instead of all the indi-viduals to match, in order to keep the completeness of explanatory evidences but reduce the computation and storage. Our previous work (Chen *et al.* 2018) presents a solution. It selects effective and important axioms according to the prediction task, and adopts the individuals involved in such axioms. Another challenge is ensuring accuracy in matching, which is under work now.

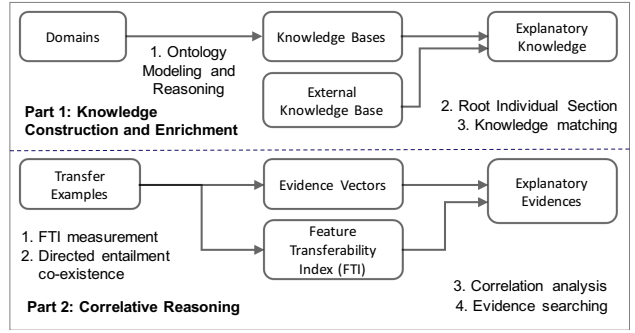

Figure 1: The Computation Framework for Explanatory Evidences for Feature Transferability

Correlative reasoning is an algorithm that infers the correlation between a feature's transferability and a candidate evidence's co-existence in source and target domains from a set of given transfers. An axiom or axiom combination is selected as a positive (negative) evidence if the correlation coefficient exceeds a positive (negative) threshold and is significant in statistic testing.

In correlative reasoning, one challenge is measuring the transferability of a feature. We consider the feature's sepcificity and generality according to (Yosinski *et al.* 2014). Another challenge is searching axiom combinations for explanatory evidences. We search from low dimension to high dimension, and propose an early stop mechanism and fast extension mechansim for acceleration. More details of correlation reasoning and our solutions to the challenges are presented in (Chen *et al.* 2018).

With such human understandable explanations to the transferability of CNN features, none ML experts can optimize a transfer learning procedure by *(i)* finding out when to transfer (i.e., at which convolutional layer) for given source and target domains, and *(ii)* selecting what source domains to transfer for a given target domain. However, the evaluation of explanatory evidences, such as their generalization (i.e., performance in applying evidences inferred from one set of transfers to another set of transfer) and quality, is still an open problem.

**Zero-shot Learning Justification**

Zero-shot learning (ZSL) is to categorize samples of new classes (i.e., classes unseen in the training data) by transferring features from seen classes (i.e., classes of the training data) without any additional sample annotation effort (Palatucci *et al.* 2009). However, the transferring process is often complex, vague and uninterpretable. According to our knowledge, there are currently few work that can provide evidences to justify such a process, especially towards people

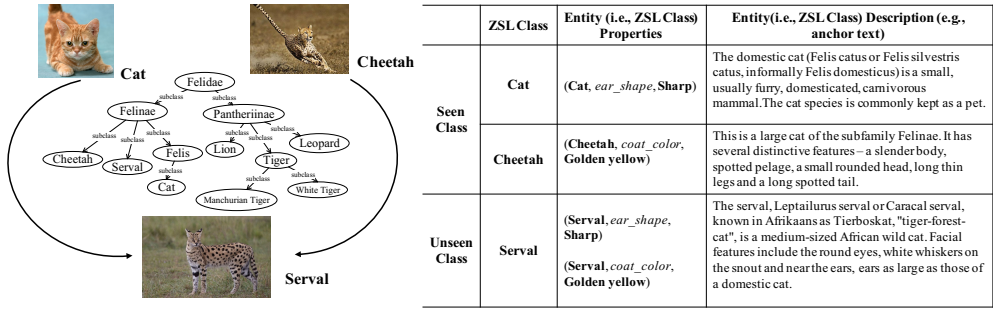

Figure 2: An example of KG-based ZSL, where *Serval*, a kind of animal with a cat-like face and a cheetah-like body is an unseen class, while *Cat* and *Cheetah* are two of the seen classes. A ZSL algorithm is to automatically generate the classification model for *Serval*, by transferring features from other seen classes using entity relations (graph structure), entity properties and descriptions of a KG (e.g., (Wang *et al.* 2018; Kampffmeyer *et al.* 2018)). Each class in ZSL is equivalent to an entity in KG.

without ML expertise.

An example of KG-based ZSL classification is shown in Figure 2, where testing samples of an unseen class *Serval* is to be classified. We first assume that a ZSL method builds a model for *Serval* by transferring the feature knowledge from a set of seen classes, according to their internally implicit and close relationships in semantics of class (KG entity) name, description, property and hierarchical structure. The target of our X-ZSL method is to find out some evidences to justify why the generated prediction model for the unseen class *Serval* is reasonable, with the help of common sense and background knowledge of these classes. One example to justify this model is "the two seen classes *Cat* and *Cheetah* play an import role in transferring features to *Serval*, while they share the same ancestors *Felinae*." It utilizes the *subClassOf* relation between entities in the KG. Another justification example is "the two seen classes *Cat* and *Cheetah* transfer their local features to the visual features of *Serval*, as the ear shape of *Serval* and *Cat* are both *Sharp*, and *Serval* has the same *Golden yellow* coat as *Cheetah*." These evidences are supported by the shared properties of *Sharp* and *Golden yellow* in the KG.

**Technical Framework**

Our ZSL justification framework includes two modules: a KG-based ZSL classifier and an justification generator, as shown in Figure 3. The classifier learns to predict the model of unseen classes with a Graph Attention Network (Velickovic *et al.* 2017). It is able to learn the attentions of the seen classes, from which we can select impressive ones. The explanation generator extracts convincing justifications from the KG according to the impressive seen classes, using class name to entity matching, SPARQL queries and reasoning.

**KG-based ZSL Classifier** We first build a hierarchy structure among all classes via knowledge graph, where each entity models one class. With the graph input, we use a kind of Graph Neural Network called Graph Attention Network (GAT) (Velickovic *et al.* 2017) to transfer the features from different unseen classes (entities) to each seen class with convolutional layers and attention layers. With the transfer, we eventually use logistic regression to predict a CNN classifier for an unseen class (i.e., the combination parameters of pre-trained CNN features).

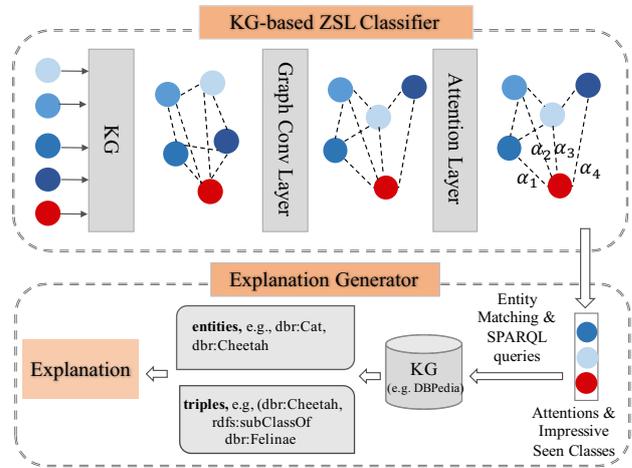

Figure 3: The technical framework for ZSL justification. The KG-based ZSL classifier learns the attention weight of seen classes ($\alpha_i$) w.r.t. an unseen class. Those with high attentions are selected as justification evidences. The explanation generator utilizes the KG (e.g., DBPedia) to obtain common sense and background knowledge to enrich the justification evidences so that common people can understand.

For each convolutional operation, GAT leverages a self-attention mechanism to update the features distribution of entities in the graph, where one entity includes features from its neighborhoods. So we can assign different weights to different entities in a neighborhoods. We also realize that not all seen classes contributes equally to the transferring to an unseen class. Thus the attention-weighted seen classes in the transferring process are extracted. Those seen classes with high attention weights in transferring are called *impressive classes*. They are selected as evidences for justification.

Current KG-based ZSL methods utilize the graph structure and class name (Wang *et al.* 2018; Kampffmeyer *et al.*

2018), but ignore the semantics from properties. To fully utilize the semantics of a KG, we transform some entity properties from the KG to so called *anonymous entities*, which is a knowledge representation and reasoning technique. Besides improving the accuracy of ZSL classifier, this will provide additional justification evidences.

**Explanation Generator** Some external KGs contains rich common sense and background knowledge. For example, DBPedia includes knowledge extracted from Wikipedia, with 4.58 million entities. We utilize such KGs to enrich the extracted justification evidence so that common people can understand.

Each class name is matched with an entity of the external KG by lexical index, which is built based on entity descriptions. For example, we can use the lookup service provided by DBPedia[1]. As shown in Figure 3, *Cat* and *Cheetah* are matched the KG entity $\langle dbr:Cat \rangle$ and $\langle dbr:Cheetah \rangle$ respectively[2]. The triples composed of entities and their relationships can form a hierarchical structure among entities that paired with same relations, e.g. $\langle dbr:Cheetah, rdfs:subClassOf, dbr:Felinae \rangle$ is a single unit in hierarchical graph in Figure 2. On the other hand, the rich triples also contain some properties about entities, such as the *Entity Property* column in the table of Figure 2.

We then access more **entities** and **triples** of the external KG, that are related to the impressive seen classes (justification evidences), using an SQL-like query language for RDF called SPARQL which can support knowledge reasoning. With returned entities and triples, we enrich the evidence to justify the generation of the classifier of each unseen class, during which some hand-crafted templates may be needed. The challenge here is to select and evaluate the returned entities for high quality explanation.

Text description of the returned entities from the external KG can also be used to enrich the explanation. However, the noise of text description is a challenge to deal with.

## Conclusion and Outlook

Transfer learning is playing a more and more important role in building complex ML systems, but the methods are still uninterpretable, especially to none ML experts. In this paper, we introduce our KG-based methods to provide human-understandable evidence for feature transferability explanation and ZSL justification. For the former, we infer evidences from axioms of OWL ontologies to explain the transferability of features learned by a CNN, while for the later, we use entity relations and properties (RDF triples) to justify the classification model generated for an unseen class. Different KG techniques, including ontology modeling, semantic reasoning, knowledge matching, semantic query and so on are incorporated for high quality explanations.

Although the solutions to the two transfer learning problems are different and there are quite a few challenges to deal with, two core and common technical problems for human-centric transfer learning explanation in our mind are knowledge matching and explanation evaluation. Knowledge matching refers to matching ML data (often in tabular format) with KG e.g., classes in ZSL with entities in DBPedia, or matching one KG with another KG. It significantly impacts the quality of explanations. In the future work, we will apply a hybrid matching approach combing lexical index, semantic reasoning and deep learning (cf. our work ColNet (Chen *et al.* 2019) and BootOX/LogMap (Jiménez-Ruiz *et al.* 2015)). While for explanation evaluation, we will feed human knowledge back to the transfer learning system through the explanation.

## Acknowledgments

The work is supported by NSFC 91846204/61673338 and the SIRIUS Centre for Scalable Data Access (Research Council of Norway, project 237889).

---

[1] https://github.com/dbpedia/lookup

[2] dbr represents http://dbpedia.org/resource, a URI prefix of DBPedia.